\documentclass[letterpaper, 10 pt, conference]{ieeeconf}

\renewcommand{\baselinestretch}{0.97}

\usepackage{textcomp}

\usepackage{cite}

\usepackage{graphicx}

\usepackage{algpseudocodex}
\usepackage{algorithmicx}
\usepackage{algorithm}
\usepackage{amsmath}

\usepackage{booktabs}

\begin{document}
%

\title{\LARGE \bf Supercomputing for High-speed Avoidance and Reactive Planning in Robots}


\author{
Kieran S. Lachmansingh, Jos\'{e} R. Gonz\'{a}lez-Estrada, Jacob Chisholm, Ryan E. Grant, Matthew K. X. J. Pan\\
Ingenuity Labs Research Institute\\
Queen's University\\
Kingston, Ontario, Canada
}


\maketitle

\begin{abstract}
This paper presents SHARP (Supercomputing for High-speed Avoidance and Reactive Planning), a proof-of-concept study demonstrating how high-performance computing (HPC) can enable millisecond-scale responsiveness in robotic control. While modern robots face increasing demands for reactivity in human–robot shared workspaces, onboard processors are constrained by size, power, and cost. Offloading to HPC offers massive parallelism for trajectory planning, but its feasibility for real-time robotics remains uncertain due to network latency and jitter. We evaluate SHARP in a stress-test scenario where a 7-DOF manipulator must dodge high-speed foam projectiles. Using a hash-distributed multi-goal A* search implemented with MPI on both local and remote HPC clusters, the system achieves mean planning latencies of 22.9 ms (local) and 30.0 ms (remote, ~300 km away), with avoidance success rates of 84\% and 88\%, respectively. These results show that when round-trip latency remains within the tens-of-milliseconds regime, HPC-side computation is no longer the bottleneck, enabling avoidance well below human reaction times. The SHARP results motivate hybrid control architectures: low-level reflexes remain onboard for safety, while bursty, high-throughput planning tasks are offloaded to HPC for scalability. By reporting per-stage timing and success rates, this study provides a reproducible template for assessing real-time feasibility of HPC-driven robotics. Collectively, SHARP reframes HPC offloading as a viable pathway toward dependable, reactive robots in dynamic environments.


\end{abstract}


%
\IEEEpeerreviewmaketitle

\section{Introduction}




Modern robots are increasingly expected to operate in unstructured and dynamic environments, often in close collaboration with humans. These scenarios demand not only accurate planning but also highly reactive control: a robot must perceive changes and adapt within tens of milliseconds to avoid unsafe or undesirable interactions. Traditionally, such responsiveness has relied on local computing platforms (commodity CPUs or GPUs embedded on the robot or on nearby workstations), which are constrained by size, power, and cost \cite{AliOnRobots,Neuman2022TinyRobots}. As robots integrate more resource-intensive artificial intelligence (AI) and machine learning models, these platforms are reaching their limits, creating bottlenecks that threaten truly responsive human–robot interaction (HRI).



High-performance computing (HPC), or supercomputing, offers an alternative. HPC systems perform billions of computations per second and excel at large-scale graph search and optimization—problems central to robotics tasks such as inverse kinematics, trajectory planning, and collision avoidance \cite{Rakita2019STAMPEDE:Kinematics,Natarajan2024PINSAT:Planning}. In principle, HPC can deliver solutions orders of magnitude faster than local compute, enabling behaviours that are more optimal, adaptive, and reactive.


The main obstacle is latency \cite{Seisa2025Cloud-AssistedImplementation}. HPC resources are typically non-local and accessed via network connections that introduce delays. In time-critical robotics, where milliseconds matter, the central question is whether the raw compute advantage of HPC can overcome communication costs and still enable real-time action \cite{IchnowskiFogROS2:2,Tahir2025EdgeSurvey}.

This paper addresses that question through \textbf{SHARP}—Supercomputing for High-speed Avoidance and Reactive Planning—a proof-of-concept system that evaluates whether HPC offloading can support real-time avoidance. To stress-test responsiveness, we examine a deliberately stringent scenario: a 7-DOF manipulator tasked with dodging high-speed foam projectiles. This setup serves two purposes. First, it creates a controlled, time-critical environment in which HPC planning must deliver trajectories fast enough to influence live robot behaviour. Second, it provides a proxy for broader applications such as collision avoidance in dynamic, human–robot shared workspaces, where comparable millisecond-scale reaction budgets apply.





\begin{figure}
\centering
\fbox{\includegraphics[scale=0.28]{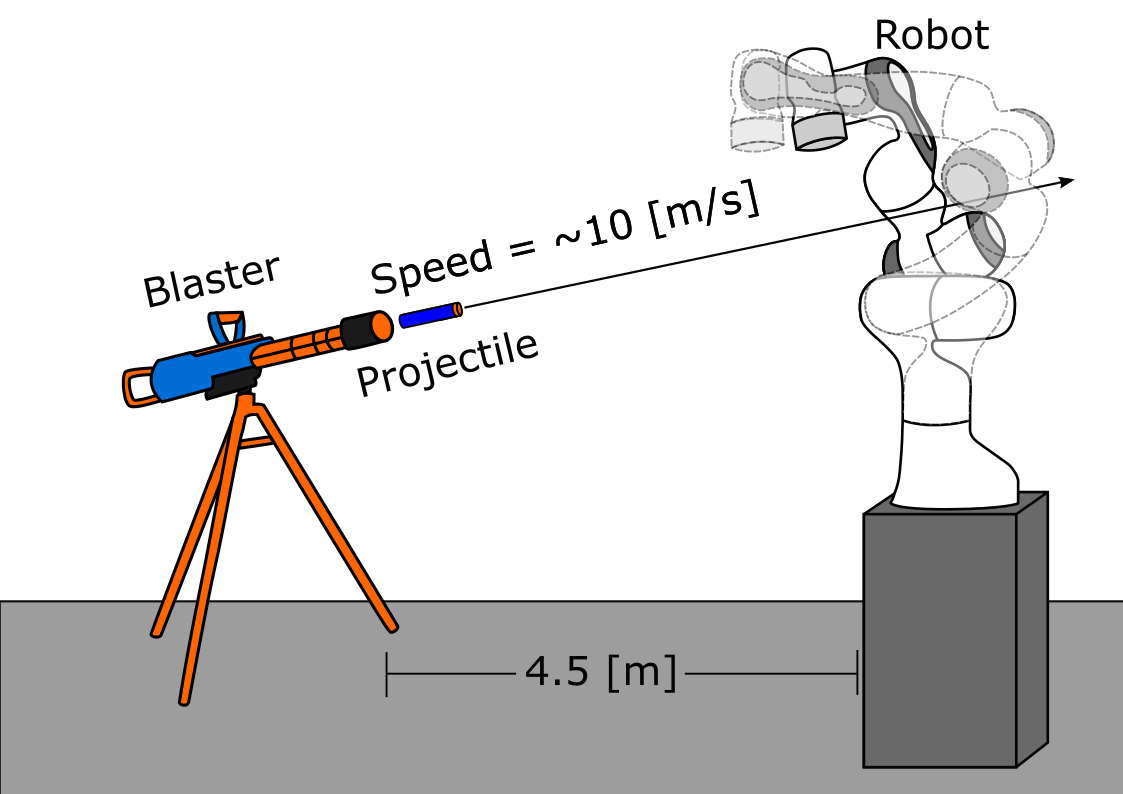}}

\caption{Projectile-dodging benchmark: The dotted outlines denote the pre-avoidance state; the solid model shows the executed post-dodge robot configuration and the measured dart pose used for prediction.}
\label{test_environment}
\end{figure}
\section{Background}
\label{sec:background}

\subsection{High-Performance and Cloud Computing in Robotics}
Robotics research has increasingly explored HPC to overcome the limitations of onboard processors for demanding tasks as robots often cannot carry high-end CPUs/GPUs needed for demanding algorithms \cite{Chen2021FogROS:Deployment}. `Cloud robotics' paradigms offers an alternative by offloading computation to clusters, effectively giving robots access to scalable processing power \cite{Camargo-Forero2018TowardsComputing}. Early frameworks like DAvinCi and Rapyuta exemplified this approach: DAvinCi integrated a Hadoop-based cluster with ROS to perform computationally intensive jobs (e.g. SLAM) in the cloud \cite{Arumugam2010DAvinCi:Robots}, while Rapyuta allowed robots to offload tasks and share a cloud-based knowledge repository (RoboEarth) to augment their local intelligence\cite{Mohanarajah2015Rapyuta:Platform}. Since then, cloud-based HPC has supported core robotic tasks, often with significant performance gains \cite{Wan2016CloudIssues}. 


However, off-site HPC introduces challenges. Network latency undermines real-time autonomy, and concerns remain about security and reliance on external infrastructure \cite{Du2017RobotComputing}.
Recent frameworks like FogROS and FogROS2 address these concerns by allowing selective offloading: developers can designate which ROS nodes run locally versus in the cloud \cite{Chen2021FogROS:Deployment}, \cite{IchnowskiFogROS2:2}. In this architecture, a robot can offload heavy components (like SLAM, grasp planning, or motion planning) to cloud GPUs/CPUs. This hybrid approach has achieved order-of-magnitude speedups with modest overhead (0.5–1.2~s latency for 4–31× faster computation). This demonstrates that HPC resources (whether remote data centers or local clusters) can dramatically accelerate robotic algorithms, as long as the system is designed to tolerate the communication delays. 


\subsection{HPC and Cloud-based Planning for Robot Obstacle Avoidance}
Trajectory planning and obstacle avoidance in complex environments remain computationally intensive, motivating HPC-based methods; traditional single-CPU planners often struggle to meet real-time requirements when the search space or number of obstacles grows large\cite{Abuelsamen2025IndustrialSystems}. Parallel sampling-based approaches mitigate this: multiple RRT or RRT* instances can run concurrently, halting when the first finds a path \cite{Kuffner2000RRT-connect:Planning}. Ichnowski et al. extend this to cloud serverless computing, launching many parallel planners as independent functions that internally exploit multi-core CPUs, enabling interactive planning for high-dimensional tasks \cite{Ichnowski2020FogComputing}. This multi-level parallelism (threads within each function, and many functions in parallel) allowed motion plans to be found much faster than a single-core baseline, effectively approaching interactive planning even for high-dimensional manipulation tasks. The ability to scale out computation on demand – even if only for bursts – is a clear advantage of HPC in time-sensitive planning: a robot facing a difficult planning problem can temporarily enlist dozens of cores in the cloud to find a path in seconds rather than minutes.

\section{Contributions} 
Despite evidence that HPC can accelerate robotic algorithms as presented in Sec. \ref{sec:background}, open questions remain about whether its raw computational speed can overcome the latency introduced by non-local access in real-time tasks. Most prior studies examine offloaded planning in offline or asynchronous contexts, with little attention to end-to-end responsiveness in dynamic, safety-critical settings. Furthermore, while latency challenges are well documented in teleoperation and cloud robotics, few frameworks investigate the feasibility of HPC for autonomous robots operating in fast, unpredictable environments.

This paper introduces SHARP, the first proof-of-concept demonstration of supercomputing applied to millisecond-scale reactive robotics. Our contributions are fourfold: 
\begin{enumerate}
    \item We design and implement a dataflow where a robot offloads avoidance requests to an HPC, receives executable trajectories, and closes the loop in real time.
    \item We adapt a deterministic, voxelized hash-distributed A* search to HPC clusters via MPI, exploiting large-scale parallelism to guarantee bounded runtimes for evasive motion planning.
    \item In a projectile-dodging stress test, SHARP achieves mean planning latencies of 22.9 ms (local) and 30.0 ms (remote, ~300 km) with avoidance success rates of 84–88\%, demonstrating viability under sub-human reaction time budgets.
    \item A template for end-to-end evaluation—reporting per-stage timing and success on viable shots—that other HPC-controlled robotics systems can adopt, providing a method to compare time-sensitive offloading.
    
\end{enumerate}

Together, these contributions provide a missing link between algorithmic parallelism and practical deployment, highlighting how HPC can extend robotic responsiveness in scenarios where timing is critical.


\section{System Overview}


\subsection{Proof-of-Concept Demonstration Description}

To evaluate the feasibility of HPC for time-sensitive robotics, we designed a deliberately stringent stress-test scenario: a 7-DOF manipulator tasked with dodging high-speed foam projectiles. While playful on the surface, this scenario imposes strict timing constraints and serves as a proxy for broader challenges in dynamic environments and human–robot shared workspaces.

The benchmark highlights two critical aspects of reactive control. First, it forces the system to close the loop—from perception, to offloading, to HPC computation, to robot actuation within a few tens of milliseconds. If the robot can successfully avoid fast-moving projectiles, it can also handle slower and less structured obstacles, such as humans entering its workspace. Second, the benchmark exposes the interplay between sensing fidelity, network overhead, and remote computation in a safety-critical context. Latency or jitter at any stage directly impacts whether avoidance succeeds.

We therefore adopt the projectile-dodging task as a stress test for SHARP. By pushing the system to operate under severe reaction-time requirements, we can quantify both its potential and its limitations. The results reveal not only whether HPC offloading can support reactive behaviour in principle, but also under what conditions it remains viable when milliseconds matter.

\subsection{Hardware}

\subsubsection{Robot} We use a Franka Emika Panda, a 7-DOF manipulator. Its redundant configuration allows the robot to perform evasive manoeuvres while maintaining a desired position and orientation of the end-effector if allowed. The robot runs under ROS2 Humble with a real-time control frequency of 1 kHz (through the Franka Control Interface), ensuring fast and reliable command execution \cite{FrankaDocumentation}.

\subsubsection{Projectile Generator} A Zuru X-Shot blaster \footnote{https://zurutoys.com/brands/x-shot/insanity} shoots foam darts at the robot. The blaster is instrumented with a Raspberry Pi Pico W, which provides strict timing triggers at each projectile launch over Bluetooth, ensuring synchronized avoidance requests.  


\subsubsection{Motion Capture} A Vicon motion capture system (12 cameras, 240 Hz) provides high-fidelity ground truth of blaster pose and projectile trajectory, isolating compute responsiveness from perception noise.  


\subsubsection{HPCs} 
\label{sec:hpcs}
 Two HPC platforms are used: (i) a local 12-node Xeon cluster with sub-ms network RTT, and (ii) a remote +1000-node EPYC cluster located $\sim$300 km away, enabling evaluation under realistic latency.

\begin{itemize}
    \item Local cluster – a small (12) multi-node HPC located locally (on campus) equipped with 2× Intel Xeon Gold 6338 CPUs (64 cores), providing a controlled environment with minimal network overhead.
    \item Remote cluster – a large (+1000) multi-node HPC located $\sim$300~km away, featuring 2× AMD EPYC 7532 CPUs (64 cores) per compute node, accessed non-locally to evaluate the impact of real-world network latency on responsiveness.
\end{itemize}

These complementary platforms allow us to compare performance in both near-ideal (local) and practical (remote) HPC configurations.


\subsection{System Workflow}





\begin{figure}[h!]
\centering
\includegraphics[scale=0.48]{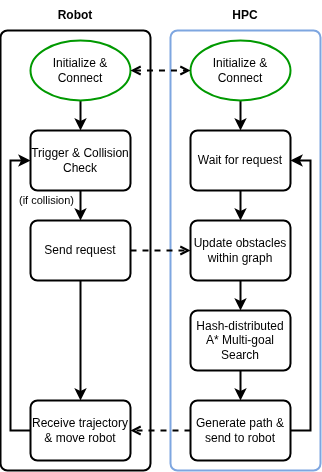}
\caption{General SHARP system overview.}
\label{sys_overview}
\end{figure}

Figure~\ref{sys_overview} summarizes the closed-loop workflow of SHARP. At runtime, the robot continuously monitors for projectile launches while maintaining a live connection to the HPC cluster. When a firing trigger is received, the system checks for potential collisions and, if needed, generates a compact voxelized representation of the predicted projectile trajectory. This representation, along with the robot’s state, is transmitted to the HPC cluster.

On the HPC side, the voxel data is distributed across processes and evaluated in parallel. A hash-distributed multi-goal A* search identifies a feasible avoidance trajectory, which is then returned to the robot in compact waypoint form. The robot immediately executes this motion using a real-time controller, before resuming its monitoring state.

This loop (sense, offload, plan, and execute) repeats continuously, enabling the robot to respond to successive projectiles in real time. Each stage of the pipeline is detailed in Sec.~\ref{sec:methodology}.

\section{Methodology}
\label{sec:methodology}

This section details the end-to-end pipeline that enables SHARP to perform HPC-driven avoidance in real time. As outlined in Fig.~\ref{sys_overview}, the system operates as a closed loop with four stages: (i) collision detection and environment encoding, (ii) parallel planning on the HPC cluster, (iii) trajectory generation and return, and (iv) execution on the robot. For clarity, we also describe how each stage’s latency is measured.


\subsection{Collision Detection and Encoding}
When a projectile is launched, a trigger from the instrumented blaster signals the robot computer to evaluate whether its current configuration is at risk of collision. Using the most recent Vicon pose of the blaster, the predicted projectile path is approximated as a straight line, sufficient to enforce strict timing constraints.

\begin{figure}[tbp]
\centering
\includegraphics[trim={5cm 5.5cm 5cm 3.5cm},clip,scale=0.34]{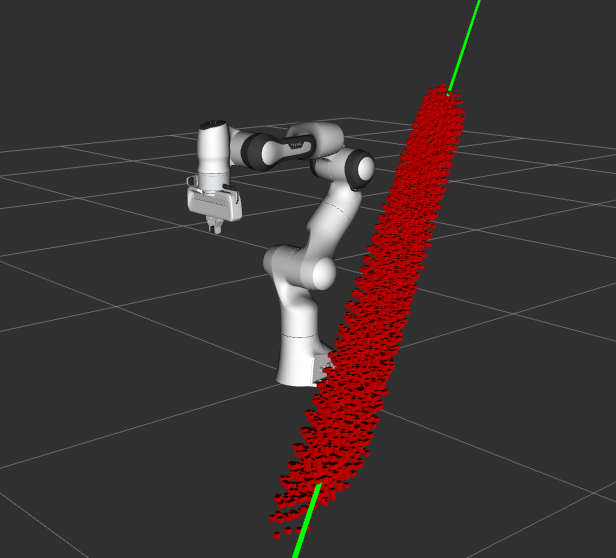}
\caption{Sample voxel generation displayed within RViz.}
\label{voxel_generation}
\end{figure}

The workspace is voxelized around the predicted trajectory (Fig.~\ref{voxel_generation}), producing a compact obstacle representation. Voxels are arranged cylindrically by height, radius, and angular resolution, with occupied voxels flagged as obstacles.

\subsubsection{Elbow Configuration Discretization}
To capture the robot’s redundancy, the null space is discretized using a custom analytic inverse kinematics solver. We define an elbow configuration as a distinct null-space solution for the 7-DOF manipulator. Each end-effector pose and orientation may therefore exist across multiple elbow configurations. For efficiency, the number of elbow configurations is fixed to a power of two. A power of two allows for easy scaling across different clusters and improved load balancing across all processes, in addition to aligning well with computer memory, improving overall performance.




\subsection{HPC Planning}
The encoded environment is transmitted to the HPC cluster for parallel trajectory planning.
\subsubsection{Graph Formulation}
The robot’s discretized workspace is modelled as a four-dimensional graph ($x,y,z,$ elbow). Nodes represent candidate Cartesian end-effector poses with associated elbow configurations. Edges represent feasible transitions, including Cartesian unit steps and diagonal connections, while disallowing discontinuous elbow jumps to force smooth trajectories. Specifically, for each node there are six edges representing Cartesian connections (unit steps along a single axis) and a further 20 edges that represent the possible diagonal motions.  


We store nodal information in four bit arrays: (1) node parents, (2) node distances from the starting node, (3) whether the node has been checked, and (4) whether it is invalid. As an example, discretizing the workspace into 1~cm$^{3}$ voxels with a 50~cm robot radius and 8 elbow configurations requires only $\sim$100~MB of storage. Increasing the resolution to 1~mm$^{3}$ voxels under the same conditions increases storage to $\sim$100~GB.


\subsubsection{Hash-Distributed Multi-Goal A*}
We employ a hash-distributed multi-goal A* search adapted for MPI. Unlike RRT-based planners, which rely on random sampling and can require many iterations to converge \cite{Kuffner2000RRT-connect:Planning}, A* is deterministic and guarantees a feasible solution within bounded time—critical for real-time avoidance. Each process expands nodes from its local heap, distributes neighbours, and synchronizes with an MPI \texttt{Allreduce} to detect the first collision-free solution. The heuristic prioritizes minimal elbow displacement before end-effector distance, keeping evasive motions compact. Algorithm~\ref{alg:multigoal} summarizes the procedure. The heuristic guiding the search is defined as:





\begin{equation}
h(s,g) = \tfrac{1}{V}\bigl(|e_s - e_g| + 0.3(|e_s - m| + |e_g - m|)\bigr)+(E(s,g))
\end{equation}

with $e_s,e_g$ the start/goal elbows, $m$ the midpoint, $V$ the elbow discretization count, and $E$ the approximate Euclidean distance. This weighting prioritizes elbow variation over end-effector displacement, minimizing deviation from the robot’s current pose.

\begin{algorithm}[]
\caption{Multi-Goal MPI A* Solver}
\label{alg:multigoal}
\begin{algorithmic}[1]
\Require Graph $G$, Source node $s$, Target node $t$, Variations $v$
\Ensure Path from $s$ to nearest valid node
\State Initialize \texttt{invalids}, \texttt{checked}, \texttt{distances}, \texttt{parents}
\State Insert $s$ into open set with $f(s) = 0$
\While{open set not empty \textbf{and} path not found}
    \State $u \gets$ extract node with minimum $f$ from open set
    \If{$u$ is valid and better than current best}
        \State Update best node and path owner
    \EndIf
    \ForAll{neighbours $n$ of $u$}
        \State $g \gets g(u) + \text{edge weight}$
        \If{$g < g(n)$}
            \State Update \texttt{parents[$n$]} $\gets u$
            \State $f(n) \gets g + h(n,t)$
            \State Insert $n$ into open set
        \EndIf
    \EndFor
    \State \textbf{MPI Allreduce:} Check if target is found across processes
    \If{a target is found}
        \State \textbf{MPI Allreduce:} Determine which process has best node
        \If{\Call{ReconstructPath}{best node, source}}
            \State \Return path
        \Else
            \State Reset search and continue
        \EndIf
    \EndIf
\EndWhile
\end{algorithmic}
\end{algorithm}


\subsubsection{Trajectory Reconstruction}
Once a valid node is found, the solver reconstructs a path of waypoints including end-effector positions, elbow configurations, and joint angles. This trajectory is returned to the robot in compact form.

\subsection{Trajectory Execution on Robot}
Upon receiving waypoints, the robot interpolates them with the Ruckig library \cite{RuckigMachines} to ensure smooth velocity and acceleration profiles. Execution proceeds under a 1 kHz proportional–derivative (PD) controller that converts tracking errors into motor torques. To reduce latency, the controller updates targets dynamically—advancing to the next waypoint once $\sim$20\% of the current path is completed. After finishing an avoidance maneuver, the robot holds its final pose and resumes monitoring for subsequent projectiles.

\subsection{Timing Breakdown}
To evaluate responsiveness, computation is partitioned into three measurable stages:
\begin{enumerate}
    \item \textbf{Unpack:} voxel decompression and synchronization across processes.
    \item \textbf{Search:} the core hash-distributed A* traversal and trajectory identification.
    \item \textbf{Backtrace:} reconstruction of waypoints for execution on the robot.
\end{enumerate}
This breakdown allows attribution of latency to communication overhead versus algorithmic complexity, a distinction that is critical for assessing the feasibility of HPC in real-time robotic control.


\section{Results} 
We evaluate SHARP along three axes: (i) \emph{responsiveness} (planning latency and its components), (ii) \emph{avoidance success rate} on valid shots, and (iii) \emph{sensitivity to communication latency} (local vs.\ remote HPC). Experiments were run on both a local (on-campus) cluster and a remote cluster (\S\ref{sec:hpcs}). We report aggregate success across all valid projectiles and per-trial rates, and decompose compute time into \emph{unpack}, \emph{search}, and \emph{backtrace} to attribute delays to communication versus algorithmic complexity (\S\ref{sec:methodology}). We also state observed limitations arising from hardware constraints and projectile variability.
\subsection{Experimental Parameters}
\label{sec:results-params}

\subsubsection{Blaster and Timing Budget}
Average projectile speed at exit was \(\sim10~\text{m/s}\) with the blaster at \(\sim 4.5~\text{m}\) from the robot, yielding a minimum time-of-flight of \(\sim 450~\text{ms}\). For context, simple human visual reaction time is \(\sim 180\text{–}200~\text{ms}\) \cite{Reaction110800}. This establishes a conservative envelope within which SHARP must complete sensing, offload, planning, return, and execution.


\subsubsection{Voxel Specification}
The workspace was discretized at \(1~\text{cm}^3\) resolution with a \(75~\text{cm}\) radius (90\% of the robot's reach), and elbow redundancy discretized to the configuration count used in \S\ref{sec:methodology}, which yielded reliable IK solutions and timely evasive motion.

\subsubsection{Hardware}

The robot manipulator was connected to a Dell G15 laptop (Ryzen 7 6800H) via Ethernet, allowing for use of the Franka Control Interface for 1kHz communication \cite{FrankaDocumentation}. For our experiments, we ran HPC computation on single compute node clusters both locally and remote (see Sec. \ref{sec:hpcs}), and on the local, robot-connected laptop (Dell G15) for comparison. The HPC computation is able to run on more compute nodes; however, for initial testing, we use only one.  Unfortunately, the laptop was unable to execute the full graph search at all due to a lack of sufficient memory and cores to run full tests.

\subsubsection{Outcome Definitions of Projectile Trajectories}

Projectiles were fired at the robot and classified as \emph{hits}, \emph{avoidances}, or \emph{invalids}. 
Hits are projectiles that contact the robot after a valid HPC request; avoidances are those that miss after a valid request; and invalids include shots without a valid request, which include those striking the immovable base of the robot, having erratic trajectories, or that are misfired. Invalids were excluded from analysis, leaving only valid shots to calculate avoidance rates. 
Because trials contained differing numbers of valid shots—due to reclassification during video review and occasional attempts to offset invalid base hits—the total varied across trials. 
This introduces a weighting effect: trials with more valid shots contribute more heavily to aggregate rates.  Accordingly, we report both per-trial success rates and aggregate success across all valid projectiles (Tables~\ref{table:orchid_shots} and \ref{table:narval_shots}).

\subsection{Local HPC Results}
Table \ref{table:orchid_shots} summarizes the results of 12 trials conducted using the local (on-campus) HPC cluster. Each trial consisted of a full sequence of projectile firings from the blaster toward the robot under fixed experimental conditions (blaster orientation, Vicon tracking setup, and HPC connection). Twelve trials were chosen to provide sufficient repetitions for averaging while keeping the experiment duration manageable. Within each trial, multiple shots were fired to capture variability in projectile speed and robot response; grouping shots into trials ensured that systematic effects (e.g., bias in path finding, cluster load) were captured across repeated runs. Bias may occur due to the robot using the same starting location throughout all trials, shown in Figure \ref{voxel_generation}. 
Across all valid shots, the robot achieved an average avoidance success rate of \textbf{84\%}, demonstrating that the system can—but not perfectly—react and avoid dynamic projectiles. Post-experiment analysis confirmed that hits occurred either due to misalignment between the real and virtual blaster or because the robot was unable to react in time due to motor speed limitations.
\begin{table}[!ht]
    \caption{The number of projectile collisions with the robot (hits), avoidances, invalids, total projectiles, and total valid projectiles within testing while using local HPC.}
    \label{table:orchid_shots}
    \centering
    \begin{tabular}{lcccccr}
    \toprule
        \textbf{Trial} & \textbf{Hits} & \textbf{Avoidance} & \textbf{Invalid} & \textbf{Total} & \textbf{Valid} & \textbf{Avoidance} \\ 
         & & & & & \textbf{Total} & \textbf{Rate}\\\midrule
        L1 & 0 & 11 & 2 & 13 & 11 & 100\% \\ 
        L2 & 2 & 12 & 7 & 21 & 14 & 86\% \\ 
        L3 & 2 & 13 & 1 & 16 & 15 & 87\% \\ 
        L4 & 1 & 12 & 5 & 18 & 13 & 92\% \\ 
        L5 & 3 & 10 & 3 & 16 & 13 & 77\% \\ 
        L6 & 2 & 4 & 13 & 19 & 6 & 67\% \\ 
        L7 & 0 & 11 & 10 & 21 & 11 & 100\% \\ 
        L8 & 1 & 8 & 10 & 19 & 9 & 89\% \\ 
        L9 & 5 & 13 & 7 & 25 & 18 & 72\% \\ 
        L10 & 0 & 13 & 6 & 19 & 13 & 100\% \\ 
        L11 & 4 & 15 & 4 & 23 & 19 & 79\% \\ 
        L12 & 3 & 4 & 9 & 16 & 7 & 57\% \\
        \textbf{Total} & \textbf{23} & \textbf{126} & \textbf{77} & \textbf{226} & \textbf{149} & \textbf{85\%} \\ 
        \textbf{Avg} & \textbf{1.9} & \textbf{10.5} & \textbf{6.4} & \textbf{18.8} & \textbf{12.4} & \textbf{84\%} \\ 
        \bottomrule
    \end{tabular}
\end{table}

Performance varied across trials, largely due to inconsistent projectile speed and blaster alignment errors. Alignment errors arose from mismatches between the simulated Vicon system and the real setup, causing deviations from the intended aim point. In particular, L6 and L12 showed elevated invalid shot counts from misfires or poor alignment, which lowered their individual success rates. These inconsistencies made trajectory prediction more difficult and introduced additional uncertainty into the avoidance task. Another limitation stemmed from the simplified projectile model. While the system assumed a linear trajectory, the foam darts frequently followed parabolic or erratic paths, reducing prediction accuracy. This mismatch highlights the need for a more realistic flight model or real-time projectile tracking to further improve dodge consistency under variable speeds.

Table \ref{table:cluster_orchid_timing} presents the compute timings for local cluster runs. Local, round-trip cluster connections used throughout the trials were measured at less than 1 ms. On average, the end-to-end compute loop required 22.9 ms, well below human visual reaction thresholds ($\sim$200 ms) \cite{Reaction110800}. The majority of this time was spent in the search phase ($\sim$20 ms), with voxel unpacking averaging just over 2 ms and backtracing under 0.2 ms. Only one trial (L5) experienced a notable spike ($\sim$49.8 ms), but even this outlier remained substantially faster than human reflexes.

\begin{table}[]
\centering
\caption{Average time per trial to unpack, search and perform a backtrace per request from the robot using local HPC.}
\label{table:cluster_orchid_timing}
\begin{tabular}{ccccc}
\toprule
\textbf{Trial}          & \textbf{Avg Unpack} & \textbf{Avg Search } & \textbf{Avg Backtrace} & \textbf{Avg Total} \\
  & \textbf{{[}ms{]}} & \textbf{{[}ms{]}} & \textbf{{[}ms{]}} & \textbf{{[}ms{]}} \\
  \midrule
L1              & 2.210                              & 22.712                  & 0.144                        & 25.587                 \\ 
L2              & 2.209                              & 13.127                  & 0.111                        & 15.940                 \\ 
L3              & 2.196                              & 14.010                  & 0.114                        & 16.828                 \\ 
L4              & 2.191                              & 12.827                  & 0.109                        & 15.644                 \\ 
L5              & 2.168                              & 46.936                  & 0.157                        & 49.760                 \\ 
L6              & 2.251                              & 14.523                  & 0.117                        & 17.419                 \\ 
L7              & 2.205                              & 16.161                  & 0.128                        & 18.969                 \\ 
L8              & 2.197                              & 13.427                  & 0.116                        & 16.260                 \\ 
L9              & 2.176                              & 20.731                  & 0.133                        & 23.561                 \\ 
L10             & 2.203                              & 17.859                  & 0.144                        & 20.717                 \\ 
L11             & 2.197                              & 22.630                  & 0.144                        & 25.471                 \\ 
L12             & 2.320                              & 24.776                  & 0.149                        & 27.772                 \\ 
\textbf{Avg} & \textbf{2.206}                     & \textbf{20.029}         & \textbf{0.131}               & \textbf{22.874}        \\ \bottomrule
\end{tabular}
\end{table}

\subsection{Remote HPC Results}
Table \ref{table:narval_shots} summarises projectile outcomes for tests executed on the remote cluster. Across all viable shots, the mean avoidance rate was \textbf{88\%}. As with local HPC results, variability across trials was influenced by invalid shots (e.g., misfires and alignment issues), which reduced the number of viable attempts in several tests.

\begin{table}[!ht]
    \centering
    \caption{The number of projectile collisions with the robot (hits), avoidances, invalids, total projectiles, and total valid projectiles within testing while using remote compute.}
    \label{table:narval_shots}
    \begin{tabular}{lcccccr}
    \toprule
    \textbf{Trial} & \textbf{Hits} & \textbf{Avoidance} & \textbf{Invalid} & \textbf{Total} & \textbf{Valid } & \textbf{Avoidance} \\ 
     & & & & & \textbf{Total} & \textbf{Rate}\\ 
    \midrule
    R1 & 1 & 14 & 1 & 16 & 15 & 93\% \\ 
    R2 & 2 & 17 & 2 & 21 & 19 & 89\% \\ 
    R3 & 2 & 8 & 9 & 19 & 10 & 80\% \\ 
    R4 & 2 & 7 & 12 & 21 & 9 & 78\% \\ 
    R5 & 1 & 8 & 14 & 23 & 9 & 89\% \\ 
    R6 & 3 & 11 & 10 & 24 & 14 & 79\% \\ 
    R7 & 1 & 11 & 10 & 22 & 12 & 92\% \\ 
    R8 & 0 & 16 & 7 & 23 & 16 & 100\% \\ 
    R9 & 0 & 9 & 12 & 21 & 9 & 100\% \\ 
    R10 & 2 & 10 & 8 & 20 & 12 & 83\% \\ 
    R11 & 0 & 13 & 10 & 23 & 13 & 100\% \\ 
    R12 & 2 & 7 & 14 & 23 & 9 & 78\% \\ 
    \textbf{Total} & \textbf{16} & \textbf{131} & \textbf{109} & \textbf{256} & \textbf{147} & \textbf{89\%} \\ 
    \textbf{Avg} & \textbf{1.3} & \textbf{10.9} & \textbf{9.1} & \textbf{21.3} & \textbf{12.3} & \textbf{88\%} \\ \bottomrule
    \end{tabular}
\end{table}
Timing measurements for the remote cluster are reported in Table \ref{table:cluster_narval_timing}. Round-trip connection latency to a short-distance ($\sim$ 300 km) remote cluster averaged 16 ms. The round-trip connection to a different long-distance remote cluster ($\sim$ 4000 km) averaged 57 ms. From this point on, the use of a remote cluster refers to the short-distance cluster. Average per-request totals were \textbf{29.98\,ms}, decomposed into unpacking (\textbf{4.01\,ms}), search (\textbf{24.39\,ms}), and backtrace (\textbf{0.14\,ms}). Compared to local compute, unpacking increased by approximately 2\,ms on average, the search phase increased by roughly 4--5\,ms, and backtrace remained similar. Occasional longer searches were observed (e.g., R4 and R6), where the average search time exceeded 40\,ms.

\begin{table}[!ht]
    \centering
    \caption{Average time per trial to unpack, search and perform a backtrace per request from the robot using remote compute.}
    \label{table:cluster_narval_timing}
    \begin{tabular}{ccccc}
    \toprule
\textbf{Trial}          & \textbf{Avg Unpack} & \textbf{Avg Search } & \textbf{Avg Backtrace} & \textbf{Avg Total} \\
  & \textbf{{[}ms{]}} & \textbf{{[}ms{]}} & \textbf{{[}ms{]}} & \textbf{{[}ms{]}} \\
  \midrule
        R1 & 4.102 & 19.722 & 0.131 & 24.882 \\ 
        R2 & 4.120 & 20.308 & 0.132 & 27.758 \\ 
        R3 & 4.065 & 10.270 & 0.113 & 15.464 \\ 
        R4 & 4.095 & 44.672 & 0.156 & 49.906 \\ 
        R5 & 4.097 & 18.393 & 0.122 & 23.528 \\ 
        R6 & 3.922 & 48.199 & 0.122 & 53.190 \\ 
        R7 & 3.949 & 17.991 & 0.142 & 22.975 \\ 
        R8 & 3.764 & 17.466 & 0.138 & 22.271 \\ 
        R9 & 3.870 & 24.374 & 0.149 & 29.370 \\ 
        R10 & 3.772 & 32.185 & 0.179 & 40.618 \\ 
        R11 & 3.984 & 16.736 & 0.115 & 21.771 \\ 
        R12 & 4.288 & 20.952 & 0.142 & 26.329 \\ 
        \textbf{Avg} & \textbf{3.993} & \textbf{24.466} & \textbf{0.136} & \textbf{30.041} \\ \bottomrule
    \end{tabular}

\end{table}

\subsection{Key Findings and Practical Envelope}
\label{sec:key-findings}

We find that HPC planning fits comfortably within reaction budgets.
Mean planning times were \(22.9~\text{ms}\) (local) and \(30.0~\text{ms}\) (remote), with rare spikes \(< 55~\text{ms}\). Even accounting for short-haul RTT (\(\sim 16~\text{ms}\)), SHARP leaves substantial time for robot execution relative to \(\sim 450~\text{ms}\) dart flight and \(\sim 200~\text{ms}\) human reaction.

Another key finding is that projectile avoidance success appears to be robust across local/remote HPC environments. Aggregate valid-shot success was \(85\%\) (local) vs.\ \(89\%\) (remote). Misses primarily reflect physical factors (dart variability, mocap alignment) and execution limits, not compute time.

Lastly, it appears that latency sensitivity is dominated by network RTT, not unpack/backtrace.
Across both clusters, \emph{search} dominates compute, while \emph{unpack} and \emph{backtrace} are negligible. This suggests further planner speedups yield diminishing returns unless perception or network RTT/jitter are reduced.

\subsection{Limitations}
Our evaluation focused on compute-side performance and did not capture the full end-to-end loop, which would also include sensing delays, message serialization, network jitter, and controller execution. The use of a motion capture system provided unusually clean state information compared to typical onboard sensors, reducing noise but limiting generalizability. In addition, the workspace and obstacle density were modest, keeping the problem size well within what a single HPC node could handle. Finally, our projectile model assumed straight-line trajectories, while the foam darts often followed parabolic or erratic paths, reducing prediction accuracy.

These factors should temper over-generalization of the reported success rates, but they do not affect the central finding: HPC planning consistently completed in the tens-of-milliseconds range, enabling high avoidance success on valid shots.

\section{Discussion}

\subsection{Compute Perfomrnace and Latency in Context}

Across local and short-haul remote trials, HPC planning completed in the tens of milliseconds, with \emph{search} dominating and \emph{unpack/backtrace} negligible (Tables~\ref{table:cluster_orchid_timing}, \ref{table:cluster_narval_timing}). Projectile avoidance success on valid shots was similar for both clusters, averaging $84\%$ and $88\%$, respectively (Tables~\ref{table:orchid_shots}, \ref{table:narval_shots}). This indicates that, once an avoidance request is formed, HPC-side computation fits comfortably within typical reaction budgets, even when examined in context with other delays (e.g., sensing, message (de)serialization, cluster scheduling), leaving time for execution. The main exogenous factor is network RTT: sub-ms locally vs.\ $\sim$16\,ms short-haul (and $\sim$60\,ms intercontinental for reference). Our results suggest offloading remains viable while RTT stays in the tens-of-milliseconds regime, consistent with wired industrial Ethernet and emerging low-jitter wireless (5G/6G) \cite{Li2025ExploringNetwork}, and well within a \(\sim\)200\,ms human reaction envelope \cite{Reaction110800}.

    

\subsection{Sources of Variability}

Outcome variability stems primarily from the physical testbed rather than compute: dart speed fluctuations, small mocap alignment errors, and a straight-line flight assumption that underfits parabolic/erratic darts. These factors raise invalids and make borderline cases harder, even when plans return quickly. Occasional search spikes align with hard grid regions (tight passages with many nodes at similar admissible $f$), not with networking or synchronization; the low variance of \emph{unpack/backtrace} supports this view.

\subsection{Key Insights}
Based on the above discussion, we reiterate the contributions of our work. First, we demonstrate an HPC-as-a-service dataflow for robotics, implemented end-to-end with per-stage timing to localize latency. Second, we provide empirical evidence of responsiveness: compute consistently remained under 50\,ms even in worst-case spikes, while avoidance success held between 84--88\% across both local and remote clusters. Third, we adapt a deterministic multi-goal A* to a 4D workspace with elbow redundancy, showing that bounded runtimes can be achieved through parallel search and that occasional spikes are linked to workspace complexity rather than system noise. Finally, we introduce a simple evaluation template—reporting unpack, search, and backtrace alongside success rates—that can be replicated by other HPC-driven robotic systems.

Taken together, these contributions show that: (1) HPC planning consistently completes in the tens-of-milliseconds range, leaving substantial budget for execution even under remote conditions. (2) Avoidance success rates above 80\% across both local and remote trials indicate robustness despite variability in projectile flight and sensing alignment. (3) Communication, not computation, is the dominant bottleneck; once RTT exceeds tens of milliseconds, responsiveness degrades regardless of algorithmic speed. These insights provide concrete evidence that HPC offloading is not “too slow” for reactive robotics, but instead represents a viable pathway when problem structure is parallelizable and network conditions are stable.




\subsection {Practical Deployments}
While our proof-of-concept shows that HPC can deliver sub-human reaction times in dynamic avoidance tasks, questions remain about deployment in real-world environments and HRI contexts. 

At the scale of our demonstration ($\sim$3~GB), a single compute node was sufficient. The benefits of HPC become more pronounced as problem sizes grow: larger voxel grids, more elbow configurations, or fleets of collaborative robots quickly exceed onboard compute limits. Mobile platforms already encounter this during computationally intensive tasks such as SLAM, grasp planning, or high-dimensional motion planning. Compared to FogROS2, which reduced motion planning to 1.2 seconds on 96 cores, our millisecond-level results on 64 cores suggest that HPC could enable real-time SLAM and similar workloads \cite{IchnowskiFogROS2:2}.

In more complex scenarios, HPC parallelism scales naturally, allowing multi-robot systems to coordinate trajectories in dynamic environments without sacrificing responsiveness \cite{Camargo-Forero2018TowardsComputing}. Established programming models such as MPI make it possible to expand planning and perception workloads without redesigning algorithms, and even the smallest system on the TOP500 list provides over 8,000 cores—far beyond the requirements of typical robotics workloads \cite{TOP500TOP500}.

Thus, the feasibility demonstrated on a single manipulator extends to broader applications. In factories, centralized clusters could coordinate fleets of cobots while reducing per-robot hardware costs. In healthcare, shared HPC services could enable safe navigation for multiple service robots operating around staff and patients. More generally, any setting that demands bursty, high-throughput planning—such as batched inverse kinematics or swarm coordination—stands to benefit from this architecture.


\subsection{Future Work}
Two directions emerge from this work. First, the HPC algorithms remain largely unoptimized. Replacing global synchronizations (e.g., MPI \texttt{Allreduce}) with non-blocking communication could reduce stalls, and scaling beyond a single 64-core node will clarify how performance changes once inter-node communication dominates. Evaluating scalability at larger problem sizes is a key next step.

Second, practical deployments will require hybrid architectures. Local controllers can handle low-level stability and reflexive safety, while HPC provides global planning for faster or less predictable obstacles. Such layered control mirrors established hierarchies in robotics but extends their reach by leveraging supercomputing resources. Exploring how best to overlap local reflexes with HPC offloading remains an open question.

\subsection{Novelty Beyond Cloud Robotics}
Prior work in cloud robotics has often concluded that offloading is too slow for real-time control, or has presented performance gains in offline or asynchronous contexts. SHARP moves this debate forward by showing that the question is not whether cloud or HPC is categorically “too slow,” but rather how parallelism and network latency interact to define a quantitative reaction-time envelope. By reporting planning latencies in the tens of milliseconds, we demonstrate that HPC offloading can meet budgets well below human reaction thresholds when round-trip times remain within the same order of magnitude. Importantly, our use of a deterministic, MPI-based multi-goal A* contrasts with earlier sampling-based approaches (e.g., RRT), providing bounded runtimes and more predictable behaviour in safety-critical settings.

\subsection{Practical Relevance to Human–Robot Interaction}
Although projectile dodging may appear stylized, we view it as a deliberate stress test. If the system can react to foam darts traveling at $\sim$10 m/s, it is over-provisioned for slower, more structured motions typical of human co-workers or mobile obstacles. This links directly to industrial cobots that must avoid encroaching workers, or hospital service robots that must maneuver safely around staff and patients. Our results suggest that HPC offloading can extend robotic responsiveness in such domains, where tens of milliseconds matter for both safety and trust.

\section{Conclusion}
In this paper, we introduced \textsc{SHARP}, an HPC-as-a-service system that offloads time-critical planning to a compute cluster and returns executable evasive manoeuvres in tens of milliseconds. As a proof-of-concept demonstration, we use SHARP in a dynamic avoidance task, where HPC-side compute averaged $22.9$\,ms (local) and $30.0$\,ms (remote) with additional RTT latency of $16$ \,ms. In this task, unpack/backtrace are negligible and compute is search dominant; viable-shot avoidance rates were $84\%$ (local) and $88\%$ (remote). These findings demonstrate that—when round-trip latency and jitter are in the tens-of-milliseconds regime—parallel planning on HPC can meet reaction budgets well below human visual response times, and that a deterministic, grid-aligned multi-goal A* (via MPI) provides bounded, parallelizable runtimes suited to this discretized setting.

Practically, our results indicate that HPC offloading is viable for reactive behaviour when (i) the planning subproblem is highly parallel (e.g., batched IK/collision checks, multi-goal grid search) and (ii) network RTT/jitter are predictably low; under these conditions, compute time becomes a small slice of the end-to-end budget, leaving room for sensing and control. Architecturally, the work motivates hybrid designs that keep last-ditch reflexes at the edge while reserving HPC for bursty, high-throughput search, with communication overlapped against expansion to smooth spikes. Methodologically, we argue that deterministic search with admissible heuristics offers stable, certifiable timing envelopes that are easier to distribute and reason about than stochastic sampling in tight control loops. Scientifically, the paper provides the first (to our knowledge) quantitative envelope that ties planner speedups to concrete latency budgets for \emph{live} control, reframing the cloud/HPC debate from a binary “too slow vs.\ fast enough” to a measured trade-off space defined by parallelism and network conditions. This establishes a template for end-to-end evaluation—reporting per-stage timing and success on viable shots—that other HPC-controlled robotics systems can adopt, thereby helping the community converge on comparable, auditable claims about time-sensitive offloading. Future work will quantify complete end-to-end latency under onboard perception, incorporate ballistic/drag or real-time projectile tracking, explore GPU-accelerated collision/IK, and study robustness under higher jitter and multi-robot settings. Taken together, these directions move HPC-driven planning from feasibility to dependable, real-time practice in shared human–robot workspaces.

\section*{Acknowledgement}

We would like to thank members of MItHRIL, CAESAR and Ingenuity Labs at Queen's University for their roles in supporting this work.



%
\bibliographystyle{IEEEtran}
\bibliography{references}

@article{Wan2016CloudIssues,
    title = {{Cloud robotics: Current status and open issues}},
    year = {2016},
    journal = {IEEE Access},
    author = {Wan, Jiafu and Tang, Shenglong and Yan, Hehua and Li, Di and Wang, Shiyong and Vasilakos, Athanasios V.},
    pages = {2797--2807},
    volume = {4},
    publisher = {Institute of Electrical and Electronics Engineers Inc.},
    doi = {10.1109/ACCESS.2016.2574979},
    issn = {21693536}
}

@article{Seisa2025Cloud-AssistedImplementation,
    title = {{Cloud-Assisted Remote Control for Aerial Robots: From Theory to Proof-of-Concept Implementation}},
    year = {2025},
    journal = {Proc. - 2025 IEEE 25th Int. Symp. on Cluster, Cloud and Internet Comput. Workshops, 2025},
    author = {Seisa, Achilleas Santi and Sankaranarayanan, Viswa Narayanan and Damigos, Gerasimos and Satpute, Sumeet Gajanan and Nikolakopoulos, George},
    pages = {171--176},
    publisher = {Institute of Electrical and Electronics Engineers Inc.},
    isbn = {9798331509385},
    doi = {10.1109/CCGRIDW65158.2025.00032}
}

@article{Arumugam2010DAvinCi:Robots,
    title = {{DAvinCi: A cloud computing framework for service robots}},
    year = {2010},
    journal = {Proc. - IEEE Int. Conf. on Robot. and Automat.},
    author = {Arumugam, Rajesh and Enti, Vikas Reddy and Bingbing, Liu and Xiaojun, Wu and Baskaran, Krishnamoorthy and Kong, Foong Foo and Kumar, A. Senthil and Meng, Kang Dee and Kit, Goh Wai},
    pages = {3084--3089},
    isbn = {9781424450381},
    doi = {10.1109/ROBOT.2010.5509469},
    issn = {10504729}
}

@article{Tahir2025EdgeSurvey,
    title = {{Edge Computing and its Application in Robotics: A Survey}},
    year = {2025},
    author = {Tahir, Nazish and Parasuraman, Ramviyas},
    month = {7},
    url = {https://arxiv.org/abs/2507.00523v1},
    arxivId = {2507.00523}
}

@article{Li2025ExploringNetwork,
    title = {{Exploring the key technologies and applications of 6G wireless communication network}},
    year = {2025},
    journal = {iScience},
    author = {Li, Pengfei and Fan, Jiaxin and Wu, Jianhong},
    number = {5},
    month = {5},
    pages = {112281},
    volume = {28},
    publisher = {Elsevier},
    doi = {10.1016/J.ISCI.2025.112281},
    issn = {2589-0042}
}

@article{Ichnowski2020FogComputing,
    title = {{Fog Robotics Algorithms for Distributed Motion Planning Using Lambda Serverless Computing}},
    year = {2020},
    journal = {Proceedings - IEEE International Conference on Robotics and Automation},
    author = {Ichnowski, Jeffrey and Lee, William and Murta, Victor and Paradis, Samuel and Alterovitz, Ron and Gonzalez, Joseph E. and Stoica, Ion and Goldberg, Ken},
    month = {5},
    pages = {4232--4238},
    publisher = {Institute of Electrical and Electronics Engineers Inc.},
    isbn = {9781728173955},
    doi = {10.1109/ICRA40945.2020.9196651},
    issn = {10504729}
}

@article{IchnowskiFogROS2:2,
    title = {{FogROS2: An Adaptive Platform for Cloud and Fog Robotics Using ROS 2}},
    author = {Ichnowski, Jeffrey and Chen, Kaiyuan and Dharmarajan, Karthik and Adebola, Simeon and Danielczuk, Michael and Mayoral-Vilches, Víctor and Jha, Nikhil and Zhan, Hugo and LLontop, Edith and Xu, Derek and Kubiatowicz, John and Stoica, Ion and Gonzalez, Joseph and Goldberg, Ken},
    url = {http://10.0.0.26/}
}

@article{Chen2021FogROS:Deployment,
    title = {{FogROS: An Adaptive Framework for Automating Fog Robotics Deployment}},
    year = {2021},
    journal = {IEEE Int. Conf. on Automat. Sci. and Eng.},
    author = {Chen, Kaiyuan Eric and Liang, Yafei and Jha, Nikhil and Ichnowski, Jeffrey and Danielczuk, Michael and Gonzalez, Joseph and Kubiatowicz, John and Goldberg, Ken},
    month = {8},
    pages = {2035--2042},
    volume = {2021-August},
    publisher = {IEEE Computer Society},
    isbn = {9781665418737},
    doi = {10.1109/CASE49439.2021.9551628},
    issn = {21618089},
    arxivId = {2108.11355}
}

@misc{FrankaDocumentation,
    title = {{Franka Control Interface Documentation}},
    url = {https://frankarobotics.github.io/docs/}
}

@article{Abuelsamen2025IndustrialSystems,
    title = {{Industrial Robot Motion Planning with GPUs: Integration of cuRobo for Extended DOF Systems}},
    year = {2025},
    author = {Abuelsamen, Luai and Student, M Eng and Lu, Ho-Wei and Tang, Wenhan and Priyadarshini, Swati and Gomes, Gabriel},
    month = {8},
    url = {https://arxiv.org/abs/2508.04146v2},
    arxivId = {2508.04146}
}

@article{AliOnRobots,
    title = {{On the Necessity of Real-Time Principles in GPU-Driven Autonomous Robots}},
    author = {Ali, Syed W and Angelopoulos, Angelos and Massey, Denver and Haddix, Sarah and Georgiev, Alexander and Goh, Joseph and Wagle, Rohan and Sarathy, Prakash and Anderson, James H and Alterovitz, Ron}
}

@article{Natarajan2024PINSAT:Planning,
    title = {{PINSAT: Parallelized Interleaving of Graph Search and Trajectory Optimization for Kinodynamic Motion Planning}},
    year = {2024},
    journal = {IEEE Int. Conf. on Intell. Robots and Syst.},
    author = {Natarajan, Ramkumar and Mukherjee, Shohin and Choset, Howie and Likhachev, Maxim},
    pages = {13907--13914},
    publisher = {Institute of Electrical and Electronics Engineers Inc.},
    isbn = {9798350377705},
    doi = {10.1109/IROS58592.2024.10801413},
    issn = {21530866},
    arxivId = {2401.08948}
}

@article{Mohanarajah2015Rapyuta:Platform,
    title = {{Rapyuta: A Cloud Robotics Platform}},
    year = {2015},
    journal = {IEEE Trans. on Automat. Sci. and Eng.},
    author = {Mohanarajah, Gajamohan and Hunziker, Dominique and D'Andrea, Raffaello and Waibel, Markus},
    number = {2},
    month = {4},
    pages = {481--493},
    volume = {12},
    publisher = {Institute of Electrical and Electronics Engineers Inc.},
    doi = {10.1109/TASE.2014.2329556},
    issn = {15455955}
}

@misc{Reaction110800,
    title = {{Reaction times to sound, light and touch - Human Homo sapiens - BNID 110800}},
    url = {https://bionumbers.hms.harvard.edu/bionumber.aspx?s=n\&v=4\&id=110800}
}

@article{Du2017RobotComputing,
    title = {{Robot Cloud: Bridging the power of robotics and cloud computing}},
    year = {2017},
    journal = {Future Gener. Comput. Syst.},
    author = {Du, Zhihui and He, Ligang and Chen, Yinong and Xiao, Yu and Gao, Peng and Wang, Tongzhou},
    month = {9},
    pages = {337--348},
    volume = {74},
    publisher = {North-Holland},
    doi = {10.1016/J.FUTURE.2016.01.002},
    issn = {0167-739X}
}

@article{Kuffner2000RRT-connect:Planning,
    title = {{RRT-connect: an efficient approach to single-query path planning}},
    year = {2000},
    journal = {Proc. - IEEE Int. Conf. on Robot. and Automat.},
    author = {Kuffner, James J. and La Valle, Steven M.},
    pages = {995--1001},
    volume = {2},
    doi = {10.1109/ROBOT.2000.844730},
    issn = {10504729}
}

@misc{RuckigMachines,
    title = {{Ruckig - Motion Generation for Robots and Machines}},
    url = {https://ruckig.com/}
}

@article{Rakita2019STAMPEDE:Kinematics,
    title = {{STAMPEDE: A discrete-optimization method for solving pathwise-inverse kinematics}},
    year = {2019},
    journal = {Proc. - IEEE Int. Conf. on Robot. and Automat.},
    author = {Rakita, Daniel and Mutlu, Bilge and Gleicher, Michael},
    month = {5},
    pages = {3507--3513},
    publisher = {Institute of Electrical and Electronics Engineers Inc.},
    isbn = {9781538660263},
    doi = {10.1109/ICRA.2019.8793617},
    issn = {10504729}
}

@article{Neuman2022TinyRobots,
    title = {{Tiny Robot Learning: Challenges and Directions for Machine Learning in Resource-Constrained Robots}},
    year = {2022},
    journal = {Proc. - IEEE Int. Conf. on Artif. Intell. Circuits and Syst., 2022},
    author = {Neuman, Sabrina M. and Plancher, Brian and Duisterhof, Bardienus P. and Krishnan, Srivatsan and Banbury, Colby and Mazumder, Mark and Prakash, Shvetank and Jabbour, Jason and Faust, Aleksandra and De Croon, Guido C.H.E. and Reddi, Vijay Janapa},
    pages = {296--299},
    publisher = {Institute of Electrical and Electronics Engineers Inc.},
    isbn = {9781665409964},
    doi = {10.1109/AICAS54282.2022.9870000},
    arxivId = {2205.05748}
}

@misc{TOP500TOP500,
    title = {{TOP500 List - June 2025 | TOP500}},
    url = {https://top500.org/lists/top500/list/2025/06/?page=1}
}

@article{Camargo-Forero2018TowardsComputing,
    title = {{Towards high performance robotic computing}},
    year = {2018},
    journal = {Robot. and Auton. Syst.},
    author = {Camargo-Forero, Leonardo and Royo, Pablo and Prats, Xavier},
    month = {9},
    pages = {167--181},
    volume = {107},
    publisher = {North-Holland},
    doi = {10.1016/J.ROBOT.2018.05.011},
    issn = {0921-8890}
}

\end{document}